\documentclass{article}

\usepackage{microtype}
\usepackage{subfigure}
\usepackage{booktabs} 
\usepackage{graphicx,amsmath,amssymb,psfrag,graphics,bm}
\usepackage{times,bbold}
\usepackage{hyperref}
\usepackage{cite}\usepackage{amsthm}\usepackage{dsfont}
\usepackage{array}\usepackage{mathrsfs}\usepackage{comment}

\usepackage{natbib}

\usepackage{algorithmic}

\usepackage[accepted]{icml2018arxiv}

\newcommand{\bw}{\text{\boldmath{$w$}}}
\newcommand{\bW}{\text{\boldmath{$W$}}}

\newcommand{\ba}{\boldsymbol{a}}
\newcommand{\bb}{\boldsymbol{b}}

\newcommand{\bI}{\boldsymbol{I}}

\DeclareMathOperator*{\argmin}{arg\,min}

\newcommand{\nn}{\nonumber}

\newcommand{\mE}{\mathbb{E}}
\newcommand{\Var}{\mathsf{Var}}

\icmltitlerunning{Layer-wise Adaptive Rate via Approximating Back-matching Propagation}

\begin{document}

\twocolumn[
\icmltitle{Train Feedfoward Neural Network with Layer-wise Adaptive Rate \\
via Approximating Back-matching Propagation}


\icmlsetsymbol{equal}{*}

\begin{icmlauthorlist}
\icmlauthor{Huishuai Zhang}{goo}
\icmlauthor{Wei Chen}{goo}
\icmlauthor{Tie-Yan Liu}{goo}
\end{icmlauthorlist}

\icmlaffiliation{goo}{Micorosoft Research Asia, Beijing, China}

\icmlcorrespondingauthor{Huishuai Zhang}{huzhang@microsoft.com}

\icmlkeywords{Machine Learning, ICML}

\vskip 0.3in
]



\printAffiliationsAndNotice{}  
\begin{abstract}
Stochastic gradient descent (SGD) has achieved great success in training deep neural network, where the gradient is  computed through back-propagation. However, the back-propagated values of different layers vary dramatically. This inconsistence of gradient magnitude across different layers renders optimization of deep neural network with a single learning rate problematic.  We introduce the back-matching propagation which computes the backward values on the layer's parameter and the input by matching backward values on the layer's output.  This leads to solving a bunch of least-squares problems, which requires high computational cost. We then reduce the back-matching propagation with  approximations and propose an algorithm that turns to be the regular SGD with a layer-wise adaptive learning rate strategy. This allows an easy implementation of our algorithm in current machine learning frameworks equipped with auto-differentiation. We apply our algorithm in training modern deep neural networks and achieve favorable results over SGD.
\end{abstract}

\section{Introduction}
Deep neural networks have been advancing the state-of-the-art performance over a number of tasks in artificial intelligence, from speech recognition \citep{hinton2012deep}, computer vision \cite{he2016deep} to natural language understanding \cite{hochreiter1997long}. These problems are typically formulated as minimizing non-convex objectives  parameterized by the neural network models. Typically the models are trained with stochastic gradient descent (SGD) or its variants and the gradient information is computed through back-propagation (BP) \cite{rumelhart1986learning}. 


However, the magnitudes of gradient components often vary significantly in neural network. Recall that one coordinate of the gradient is the directional derivative along with that coordinate which represents how a change on the weight will affect the loss rather than how we should modify the weight to minimize the loss. Thus vanilla SGD with a single learning rate could be problematic for  
the optimization of deep neural network because of the inconsistent magnitude of gradient components.  In practice, extreme small learning rate alleviates this problem but leads to slow convergence. Moreover, momentum \citep{rumelhart1986learning, qian1999momentum, nesterov2013introductory, sutskever2013importance} amends this problem by accumulating velocity along the coordinate with small magnitude but consistent direction and reducing the velocity for those coordinates with large magnitudes but opposite directions. Adaptive learning rate algorithms \citep{duchi2011adaptive, kingma2014adam} scale coordinates of the gradient by  reciprocals of some averages of their past magnitudes, confirming that weakening the affect of magnitudes of the gradient components could be favorable to the optimization from the other side. 

We want to solve this problem from another perspective. \citet{ye2017importance} suggests that the magnitude inconsistence of gradient components are mainly across layers. We can get a hint by scrutinizing the back-propagation through a fully connected layer\footnote{We omit bias terms for simplicity.} which has output $\bb$ and input $\ba$ and weight parameter $\bW$.  The layer mapping is given by $\bb = \bW \ba$. If $\delta \bb$ is the back-propagated value on the output is , i.e., the partial derivatives of the loss with respect to $\bb$, the \emph{back-propagation equation} is given by
\begin{flalign}
 &\delta a_j = \bw_{j\rightarrow}^T \delta \bb, \nn
\end{flalign}
where $\bw_{j\rightarrow}$ is the $j$-th column of $\bW$ (represents all the connections emanating from unit $j$) and $\delta a_j$ is the back-propagated value on the layer's input $a_j$ computed through back-propagation. 
If the rows of $\bW$ are initialized with unit length (roughly) to preserve the forward signal, then back-propagation through this layer would shrink the magnitude of the backward signal heavily when the number of input  is much larger than the number of output. 

We suggest a principled way to overcome the problem of magnitude inconsistence of gradient components across layers in back-propagation. Specifically, 
we compute changes $\delta'\bW$ and  $\delta' a_j$ on the weight parameter  and on the input respectively, to match the error guiding signal $\delta b_k$ as closely as possible. This motivates us to formulate the backward pass through the fully connected layer as solving a group of least-squares problems. By hiding technical details and some assumption, we propose the back-matching propagation as follows,
\begin{flalign}
&\delta' a_j = (\bw_{j\rightarrow}^T\bw_{j\rightarrow})^{-1} \left(\bw_{j\rightarrow}^T  \delta \bb\right), \label{eq:solutionaj} \\
&\delta' \bW = (\mE \ba \ba^T)^{-1}\mE [(\delta \bb) \ba^T], \label{eq:solutionW}
\end{flalign}
where the expectation is over the data points in a mini batch. 
A direct expalanation of \eqref{eq:solutionW} is that we want to change the weight matrix $\bW$ by $\delta' \bW$ to produce a desired change $\delta\bb$ on the output (or sufficiently close to) given the current input $\ba$. So can we explain equation \eqref{eq:solutionaj}. Then we use  $\delta' \bW$ to update the parameter $\bW$  and use $\delta' a_j$ as the error guiding signal to back-propagate to lower layers. 

For the back-matching propagation \eqref{eq:solutionaj}, we need to compute $(\bw_{j\rightarrow}^T\bw_{j\rightarrow})^{-1}$ which is easy since it is a scaler.  For the parameter update solution \eqref{eq:solutionW}, we need to compute an inverse  $\left(\mE \ba \ba^T\right)^{-1}$. This requires a large number of matrix inverse operations, roughly the number of neurons, and each inverse requires flops on the cubic order of the number of neurons in one layer. This hinders it to be applied to large neural networks which typically contain tens of thousands of neurons in a single layer.


Fortunately, we can work with the batch normalization (BN) technique \cite{ioffe2015batch} to circumvent this difficulty. With batch normalization, we regard $\mE \ba \ba^T$ as an identity matrix approximately and remove the inverse in \eqref{eq:solutionW}. Then with some approximation, we can reduce the back-matching propagation into a layer-wise gradient adaption strategy, which can be viewed as layer-wise adaptive learning rates when applying pure SGD. As such a layer-wise gradient adaption strategy is built upon the regular BP process, it is easy to implement in current deep learning frameworks \cite{theano, tensorflow, pytorch, CNTK}. Moreover, this strategy also works with other popular optimization techiques (momentum, ada-algorithms, weight-decay) naturally to achieve possible higher performances in machine learning tasks. We expect this layer-wise gradient adaption strategy could accelerate the training procedure. 
Surprisingly, this strategy often improves the test accuracy by a considerable margin in practice.

\subsection{Related Works}
Training neural network with layer-wise adaptive learning rate has been proposed in several previous works. Specifically, \citet{singh2015layer} suggested using $\eta \cdot (1+\log (1+1/\|\delta \bW_l\|_2))$ as the learning rate for the layer $l$.  \citet{you2017large} suggested using $\eta \cdot \frac{\|\bW_l\|_2}{\|\delta \bW_l\|_2}$ as the learning rate for layer $l$ and demonstrated that this would benefit the large-batch training.  However, the suggestion in both works mainly comes from empirical experience and do not have explanation of why the rate is set in that way. 


Our paper is related to the block-diagonal second order algorithms \citet{lafond2017diagonal, zhang2017block, grosse2016kronecker}. Specifically,  \citet{lafond2017diagonal} proposes a weight reparametrization scheme with a diagonal rescaling step-size and show its potential advantages over batch normalization. \citet{zhang2017block} proposes a block diagonal Hessian-free method to train neural networks and shows fast convergence rate over first-order methods. \citet{martens2015optimizing, grosse2016kronecker} propose the Kronecker Factored Approximation (KFA) method to approximate the natural gradient using a block-diagonal or block-tridiagonal approximation of the Fisher matrix. These second-order algorithms
 all share a layer-wise or block-diagonal structure design, which agrees with our algorithm.  However, our layer-wise adaptive learning rate comes from the perspective of back-matching propagation and is different from the second-order approximations. 

Our paper is also related to the Riemannian algorithms \cite{amari1998natural, ollivier2015riemannian, marceau2016practical}. Specifically, \citet{ollivier2015riemannian} proposes using $(\mE [\ba \ba^T m_k])^{-1}\mE [(\delta b_k) \ba^T]$ as the update for the parameter $\bw_{\rightarrow k}$, where $m_k$ is a backpropagated metric.  
Similarly, \citet{ye2017importance} advocates using  $(\mE \ba \ba^T + \lambda \bI)^{-1}\mE[(\delta \bb) \ba^T]$ as the update of the parameter. 

In comparison, the back-matching propagation comes from a different perspective that the back-propagated values should match the error guiding signal. Our layer-wise gradient adaption strategy, which is derived from back-matching propagation, is simpler than the Riemannian algorithms in terms of implementational and computational complexity. 


\section{Back-matching Propagation}
In this section, we present how the back-matching propagation works under several popular types of layers.  Specifically, we derive the formula of the backprogated values on the layer's input and on the layer's parameters given the backpropagated value on the layer's output based on the back-matching propagation.  Moreover, we  compare the back-matching propagation to the regular BP. 

We introduce several notations here (some have been used in Introduction).  Let $\ell$ denote the objective (loss function). We use $\bb$ and $\ba$ to denote the layer's output and input respectively and use parameter $\bW$ to denote the layer's parameter.  We use $\delta \bb$ to denote the back-propagated value on the layer's output. Then we use $\delta a_j$ and $\delta \bW$  to denote the back-propagated values computed through BP, and use $\delta' a_j$ and $\delta' \bW$ to denote the back-propagated values computed  through back-matching propagation. 

Let us briefly review the regular BP here. The BP propagates derivatives from the top layer back to the bottom one. Suppose we are dealing with a general layer which has forward mapping  $\bb = f(\ba; \bW)$. Then the derivative of the loss $\ell$ with respect to a specific output component  $b_k$ is $\delta b_k := -\frac{\partial \ell}{\partial b_k}$. The  \emph{BP equations} are given by
\begin{flalign}
 &\delta a_j = \sum_{k, j\rightarrow k} \frac{\partial b_k}{\partial a_j} \delta b_k \label{eq:bpequation}\\ 
 &\delta \bW=  \mE_x[ (\delta \bb(x))\ba(x)^T, \label{eq:deltaW} 
\end{flalign}
where $x$ represents a data point and  the expectation is over the data points in a mini batch. 

Next we present how the back-matching propagation back-propagated through specific layers. In order to compare conveniently, for each type of layer we first provide the BP formula and then derive the formula via back-matching propagation and in the end discuss the relation between the back-matching propagation and BP.
\subsection{Fully Connected Layer}
We first consider a fully connected layer, whose mapping function is given by\footnote{For simplicity we omit the bias term.}
\begin{flalign}
\bb = \bW \ba. 	\label{eq:fclayer}
\end{flalign}
Suppose the backpropagated values on the output are $\delta \bb$.  Following backpropagation equations \eqref{eq:bpequation} and \eqref{eq:deltaW},  we compute the backpropagated values on the input and on the weight parameter  as follows,
\begin{flalign}
 \delta& a_j(x) = \bw_{j\rightarrow}^T \delta \bb(x), \label{eq:fcbp}\\
 \delta & \bW =  \mE_x [\delta \bb(x) \ba(x)^T], \label{eq:fcdeltaW}
\end{flalign}
where $\bw_{j\rightarrow}$ is the $j$-th column of $\bW$.

We next derive the formula for back-matching propagation, where we compute $\delta' \bW$ and $\delta' a_j$ that try to match the guiding signal $\delta \bb$ as accurately as possible, in the sense of minimizing square error, 
\begin{flalign}
	&\delta' \bW  \leftarrow \argmin_{\delta'\bW} \|\delta \bb - (\delta'\bW) \ba\|_2^2 \label{eq:solveW}\\
	&\delta' \ba \leftarrow \argmin_{\delta'\ba} \|\delta \bb - \bW (\delta'\ba)\|_2^2.  \label{eq:solvebaj}
\end{flalign}
Note that by writing the matching problem as two independent problems \eqref{eq:solveW} and \eqref{eq:solvebaj}, we presume that updating $\bW$ and propagating backward values $\delta'\ba$ are independent, and such layer independence has been used in block-diagonal second-order algorithms \cite{zhang2017block, lafond2017diagonal}. Moreover, \eqref{eq:solveW} is separable along the rows of $\delta' \bW$. Hence we obtain a bunch of (total number $\#row(\bW)$) least-squares problems
\begin{flalign}
	&\delta' \bw_{\rightarrow k}  \leftarrow \argmin_{\delta'\bw_{\rightarrow k}} \|\delta \bb_k - \delta'\bw_{\rightarrow k} \ba\|_2^2, \label{eq:solveWk}
\end{flalign}
where $\bw_{\rightarrow k}$ is the $k$-th row of $\bW$ and $\delta\bb_k$ represents the back-propagated values at neuron $k$ in one mini batch of data. We further assume all $a_j(x)$ are updated independently, based on the intuition that a neuron doesn't know the other neurons' states \emph{on/off} and a fair strategy is to try its best to match the guiding signal by itself. Then \eqref{eq:solvebaj} becomes a bunch (total number $dim(\ba)$) of least-squares problems 
\begin{flalign}
	&\delta' a_j \leftarrow \argmin_{\delta a_j} \|\delta \bb - \bw_{j\rightarrow} \delta a_j\|_2^2,  \label{eq:solveaj}
\end{flalign}
where $\bw_{j\rightarrow}$ are the weights emanating from neuron $j$ (a column of $\bW$ corresponding neuron $j$). We call equations \eqref{eq:solveW} and \eqref{eq:solveaj} the back-matching propagation rule.
Solving the least-squares problems \eqref{eq:solveaj} and \eqref{eq:solveW} gives us:
\begin{flalign}
&\delta' a_j(x) =(\bw_{j\rightarrow}^T\bw_{j\rightarrow})^{-1} \left(\bw_{j\rightarrow}^T  \delta \bb(x)\right), \label{eq:bmpfcaj}\\
&\delta' \bW = (\mE_x \ba(x) \ba(x)^T)^{-1}\mE_x [\delta \bb(x) \ba(x)^T]. \label{eq:bmpfcW}
\end{flalign}

From \eqref{eq:fcbp} \eqref{eq:fcdeltaW} and \eqref{eq:bmpfcaj} \eqref{eq:bmpfcW}, we can see how the back-matching propagation is related with the regular BP:
\begin{flalign}
\delta'& a_j(x) = (\bw_{j\rightarrow}^T\bw_{j\rightarrow})^{-1} \delta a_j(x), \label{eq:fcbmp}\\
\delta'& \bW = (\mE_x \ba(x) \ba(x)^T)^{-1} \delta \bW. \label{eq:fcdeltapW}
\end{flalign}
We can see that the formulas \eqref{eq:fcbmp} \eqref{eq:fcdeltapW} of back-matching propagation are the corresponding BP formulas \eqref{eq:fcbp} \eqref{eq:fcdeltaW} rescaled by a number or a matrix. 

\subsection{Convolutional Layer}
In this section we study the back-matching propagation through a convolutional layer. The weight parameter $\bW$ is an array with dimension $n\times m \times w \times h$, where  $n$ and $m$ are the number of output features and the number of input features respectively, and $w$ and  $h$ are the width and height of convolutional kernels. We use $b_{ku_1u_2}$ to denote the output at location $(u_1, u_2)$ of feature $k$ and  $a_{ju_1u_2}$ to denote the input at location $(u_1, u_2)$ of feature $j$, then the forward process is 
\begin{flalign}
b_{ku_1u_2} = \sum_{j=1}^{n} \sum_{v_1v_2} a_{j(u_1+v_1)(u_2+v_2)} w_{jkv_1v_2}, \label{eq:convfw}
\end{flalign}
and the BP is given by
\begin{flalign}
\delta& a_{ju_1u_2} = \sum_{k=1}^{m} \sum_{v_1v_2} \delta b_{k(u_1+v_1)(u_2+v_2)} w_{jkv_1v_2}, \label{eq:convbp}\\
\delta& w_{jkv_1v_2} =  \sum_{u_1u_2} \delta b_{ku_1u_2} a_{j(u_1+v_1)(u_2+v_2)}. \label{eq:convdeltaW}
\end{flalign}
However, this formula of the forward and backward process of convolutional layer make the derivation of back-matching propagation complex. Note that the convolution operation essentially performs dot products between the convolution kernels and local regions of the input. The forward pass of a convolution layer can be formulated as one big matrix multiply with \emph{im2col} operation. In order to describe back matching process clearly, we rewrite the convolution layer forward and backward pass with \emph{im2col} operation. We use $\bW_{row}$ and $\bW_{col}$ to represent the weight matrices with dimension $n\times (mwh)$ and $m\times (nwh)$, respectively, which both are stretched out from $\bW (n, m, w, h)$. To mimic the convolutional operation, we rearrange the input features $\ba$ into a big matrix $\ba_{i2c}$ through \emph{im2col} operation: each column of $\ba_{i2c}$ is composed of the elements of $\ba$ that are used to compute one location in $\bb$. Thus if $\bb$ has dimension $n\times q_1\times q_2$, then $\ba_{i2c}$ has dimension $mwh\times q_1q_2$. Furthermore, we stack the latter two dimensions of $\bb$ into a tall vector, denoted as $\bb_{col}$ which has dimension $n\times q_1q_2$. The forward process \eqref{eq:convfw} of convolutional layer can be rewritten as
\begin{flalign}
\bb_{col} =\bW_{row} \ba_{i2c}
\end{flalign}
Similarly, we can rewrite the regular BP \eqref{eq:convbp} and \eqref{eq:convdeltaW} as
\begin{flalign}
&\delta a_{ju_1u_2}(x) = \bw_{ju_1u_2 \rightarrow}^T \delta \bb[ju_1u_2](x), \label{eq:convbpl} \\
&\delta \bW_{row} = \mE_x \delta \bb_{col}(x) \ba_{i2c}^T(x), \label{eq:convdeltaWl}
\end{flalign}
where $\bw_{ju_1u_2 \rightarrow}$  is composed of weight components that interact with input location $ju_1u_2$, which approximately has $nwh/c$ elements and $c$ is a factor related with pooling and stride, and $\delta \bb[ju_1u_2]$ is composed of output locations that have interaction with input location $ju_1u_2$. With these notations, we can derive  the formula for back-matching  propagation via solving the least squares problems  \eqref{eq:solveaj} and  \eqref{eq:solveW}, given by
\begin{flalign}
\delta' &a_{ju_1u_2}(x)  =  \frac{\bw_{ju_1u_2 \rightarrow}^T \delta \bb[ju_1u_2](x)}{\bw_{ju_1u_2 \rightarrow}^T \bw_{ju_1u_2 \rightarrow}}, \label{eq:convbmp} \\
\delta' &\bW_{row}^T = (\mE_x \ba_{i2c} \ba_{i2c}^T)^{-1} \mE_x \delta \bb_{col}(x) \ba_{i2c}^T(x). \label{eq:convdeltapW}
\end{flalign}
We can see that the formulas  \eqref{eq:convbmp} \eqref{eq:convdeltapW}  of back-matching propagation are the corresponding BP formulas \eqref{eq:convbpl} \eqref{eq:convdeltaWl} rescaled by a number or a matrix.  
As the convolutional layer is essentially a linear mapping, the formulas here is similar to those of the fully connected layer although they are more involved.

\subsection{Batch Normalization Layer}
Batch normalization (BN) is widely used for accelerating training of feedforward neural networks. In practice, BN is usually inserted right before the activation function. We fix the affine transformation of batch normalization to be identity. Then the BN layer mapping is given by
\begin{flalign}
	b_k = \text{BN}\left(a_k\right) = \frac{a_k- \mE[a_k]}{\sqrt{\Var[a_k]}}.
\end{flalign}
The BP formula through the BN layer  is given by \citep{ioffe2015batch},
\begin{flalign}
\delta a_k = \frac{\delta b_k}{\sqrt{\Var[a_k]}} + \frac{\left(\delta \Var[a_k] \cdot 2(a_k-\mE[a_k]) +\delta \mE[a_k]\right)}{m}, \label{eq:bnbp}
\end{flalign}
where $m$ is the mini-batch size, and $\delta \Var[a_k]$ and $\delta \mE[a_k]$ is the backpropagated values on quantities $\Var[a_k]$ and $\mE[a_k]$ respectively.

We next derive the formula of back-matching propagation through BN. By solving \eqref{eq:solveaj}, we have 
\begin{flalign}
\delta' a_k &= \delta b_k \cdot \sqrt{\Var[a_k]}. \label{eq:bmpbn}
\end{flalign}
To see how the back-matching propagation is related with BP, we ignore the latter two terms in \eqref{eq:bnbp} when the mini-batch size is large, and have the following approximation
\begin{flalign}
\delta' a_k &\approx \delta a_k \cdot \Var[a_k].
\end{flalign}


\subsection{Rectified Linear Unit (ReLU)}
We use $\sigma(\cdot)$ to denote the ReLU nonlinear function. Then the ReLU mapping is given by 
\begin{flalign}
	b_k = \sigma (a_k) =\begin{cases}
a_k , & \text{ if }a_k \ge 0\\
0, &  \text{ if } a_k < 0.
\end{cases}
\end{flalign}
For the formula of BP, we have
\begin{flalign}
\delta a_k = \delta b_k \sigma'(a_k) =\begin{cases}
\delta b_k , & \text{ if }a_k \ge 0\\
0, &  \text{ if } a_k < 0.
\end{cases}
\end{flalign}
Following \eqref{eq:solveaj}, we have the formula of back-matching propagation for the ReLU layer
\begin{flalign}
\delta' a_k = \delta b_k \sigma'(a_k).
\end{flalign}
Therefore the formula of back-matching propagation for ReLU is the same as that of BP,
\begin{flalign}
\delta' a_k = \delta a_k. \label{eq:bmprelu}
\end{flalign}

\section{{Layer-wise Adaptive Rate via Approximate Back-matching Propagation}}
The back-matching propagation involves large number of matrix inverse operations, which is computationally prohibited in training large neural networks. In this section we present how to approximate the back-matching propagation under certain assumption and end up with a layer-wise adaptive rate strategy based on the approximation of the back-matching propagation, which allows easy implementation in frameworks equipped with auto-differentiation.

\subsection{Approximate Back-matching Propagation via BP} \label{subsec:approx}
We firstly look at the formula of the back-matching propagated value on the weight parameter \eqref{eq:fcdeltapW} and \eqref{eq:convdeltapW}. It is the gradient scaled by an inverse of a matrix, which is prohibited for large networks. We use batch normalization to circumvent this difficulty. 

With batch normalization, we regard $\mE_x\ba \ba^T$ as identity matrix approximately. From now on, we require each intermediate layer is bonded with a batch normalization layer except the output layer. Since BN has been widely used for accelerating the training process and improving the test accuracy, this requirement does not confine us much.  Under this requirement, the back-matching propagation for the fully connected layer \eqref{eq:fcdeltapW} is approximated by,
\begin{flalign}
\delta' \bW = (\mE_x \ba(x) \ba(x)^T)^{-1} \delta \bW \approx \delta \bW, \label{eq:fcapproxW}
\end{flalign}
and the back-matching propagation for the convolutional layer \eqref{eq:convdeltapW} is approximated by,
\begin{flalign}
\delta' \bW_{row}^T = (\mE_x \ba_{i2c} \ba_{i2c}^T)^{-1} \delta \bW_{row}^T \approx \frac{1}{s} \delta \bW_{row}^T, \label{eq:convapproxW}
\end{flalign}
where $s$ is the sharing factor for the convolutional layer.

Next we consider the formula of back-matching propagated values on the input \eqref{eq:fcbmp} and \eqref{eq:convbmp}. To further reduce the complexity and develop a layer-wise adaptive learning rate strategy, we assume that $\bW$ is row homogeneous \cite{ba2016layer}, i.e., they represent the same level of information and are roughly of similar magnitude.  We define 
$$\|\bW\|_{2,\mu}^2 := \frac{1}{\#row(\bW)}\sum_{i=1}^{\#row(\bW)} \bw_i^T \bw_i,$$
where $\bw_i$ is the $i$-th row of $\bW$.
Then the $(\bw_{j\rightarrow}^T\bw_{j\rightarrow}) $ in equation \eqref{eq:solutionaj} can be approximated as   $(\bw_{j\rightarrow}^T\bw_{j\rightarrow})\approx \|\bW^T\|_{2,\mu}^2.$
Under this assumption, the back-matching propagation for the fully connected layer \eqref{eq:fcbmp} is approximated by,
\begin{flalign}
\delta' a_j(x) &= (\bw_{j\rightarrow}^T\bw_{j\rightarrow})^{-1} \delta a_j(x) \approx \frac{ \delta a_j(x)}{\|\bW^T\|_{2,\mu}^2} \label{eq:fcapproxbmp}
\end{flalign}
and the back-matching propagation for the convolutional layer \eqref{eq:convbmp} is approximated by,
\begin{flalign}
\delta' a_{ju_1u_2}(x)  
\approx \frac{\delta a_{ju_1u_2}(x)}{\|\bW_{col}\|_{2,\mu}^2/c},\label{eq:convapproxbmp}
\end{flalign}
where $c$ is a factor related with pooling, stride and padding operations. We will see a detailed example in Section \ref{subsec:lenet}.

For BN layer, we assume the weight parameter is row homogeneous and then the back-matching propagation for the BN layer \eqref{eq:bmpbn} is approximated by,
\begin{flalign}
\delta' a_k \approx \delta a_k \cdot \Var[a_k]\approx \|\bW\|_{2,\mu}^2\cdot \delta a_k. \label{eq:bnapproxbmp}
\end{flalign}

\subsection{Layer-wise Adaptive Rate Strategy}
Based on the approximations in Section~\ref{subsec:approx} we are ready to derive a layer-wise adaptive learning rate strategy.
We note that the approximate back-matching propagation gives a scaling factor for each layer's gradient if we back-propagate starting from the top layer. We set the initial factor of the output layer is $m_{out}=1$, which indicates that we regard the derivative of the loss with respect to the output of the network as the desired changes on the output to minimize the loss. 

Then starting from the top layer, we compute a backward factor $m$ for each layer through
 \begin{flalign}
  m_a \leftarrow m_b \cdot \frac{\delta a_j}{\delta' a_j},
 \end{flalign}
 where the relations of $\delta a_j$ and $\delta' a_j$ are given by \eqref{eq:fcapproxbmp}, \eqref{eq:convapproxbmp}, \eqref{eq:bnapproxbmp} and \eqref{eq:bmprelu} for fully connected layer, convolutional layer, BN layer and ReLU, respectively.
If the layer has parameter $\bW$ and gradient $\delta \bW$, then we use $\delta\bW/ m_b / s$ as the new adaptive gradient to update $\bW$, where $m_b$ is the backward factor on the output of the layer and $s$ is the sharing factor of the layer. Then $1/ m_b / s$ can be viewed as a layer-wise adaptive learning rate when using vanilla SGD. This strategy is described in Algorithm \ref{alg:bmp}.
\begin{algorithm}[tb]
   \caption{SGD with Layer-wise Adaptive Rate via Approximate Back-matching Propagation}
   \label{alg:bmp}
\begin{algorithmic}
   \STATE {\bfseries Initial:} \\
    Backward factor $m=1$, \\
    $s=1$ for fully connected, ReLU and BN layers,\\
    $s=  \text{weight-sharing factor}$ for convolutional layer
   \REPEAT
   \STATE BP from the layer's output
   \IF {layer has weight $\bW$}
   \STATE {$\delta' \bW \leftarrow \delta \bW / m / s$}
   \ENDIF
   \STATE {Calculate the ratio $\delta a_j/\delta' a_j$ according to the layer type }
   \STATE {Update $m \leftarrow m \cdot \frac{\delta a_j}{\delta' a_j}$}
   \UNTIL{bottom layer}
\end{algorithmic}
\end{algorithm}

Our algorithm can work with momentum naturally. In practice, we use the flow in Algorithm \ref{alg:bmp} to modify the gradient computed via BP. Then we apply the momentum update with the modified gradient. With the modified gradient given by Algorithm \ref{alg:bmp}, we can also apply other adaptive strategy, i.e., Adam and Adagrad, without difficulty.

Weight-decay is a widely used technique to improve the generalization of the model. Note that both the weight-decay and our algorithm are modifying the gradient of the network parameter computed through BP. In practice, we first apply the weight-decay modification and then apply our algorithm on the modified gradient, which produces better result than the other way around. 

\subsection{An Example: LeNet}\label{subsec:lenet}
\begin{figure}[ht]
\vskip 0.2in
\begin{center}
\centerline{\includegraphics[width=\columnwidth]{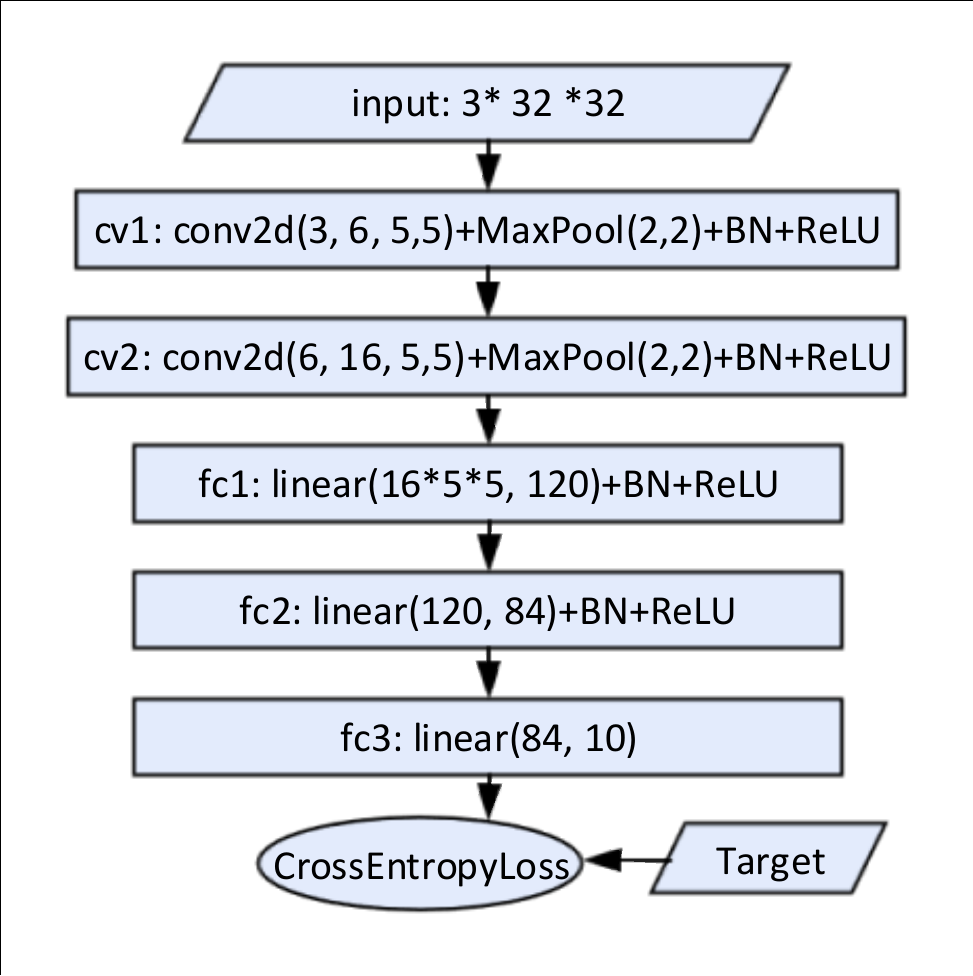}}
\caption{LeNet with batch normalization}
\label{fig:lenet}
\end{center}
\vskip -0.2in
\end{figure}

We use LeNet (Figure. \ref{fig:lenet}) as an example. We modify the original LeNet\cite{lecun1998gradient} by inserting a batch normalization transformation before each activation layer (ReLU) and omitting all the bias terms. 
We next walk through the approximated back-matching propagation of the LeNet and show how each layer's weight should be changed ($\delta'\bW$). 
Following the procedure of Algorithm \ref{alg:bmp}, we have the following initial value: $m=1, s_{fc3} = 1, s_{fc2} = 1, s_{fc1} = 1, s_{cv2} = 25, s_{cv1} = 196$. Given the loss $\ell(x)$, we can compute the normal gradient on each weight parameter through BP, which are denoted as $\delta \bW$ with subscript of the layer name. We start from the top layer \emph{fc3} and compute 
\begin{flalign*}
\delta'\bW_{fc3} = \delta \bW_{fc3} / m / s_{fc3} = \delta \bW
\end{flalign*} 
and update $m \leftarrow m\cdot \delta a_j/\delta' a_j = \|\bW_{fc3}^T\|_{2,\mu}^2$. Since the ReLU activation does not contain parameter and does not change the backward factor $m$, then we move to the BN layer. Since our BN layer does not have parameter, we only have to update the backward factor $$m \leftarrow m\cdot \delta a_j/\delta' a_j  =\frac{\|\bW_{fc3}^T\|_{2,\mu}^2}{\|\bW_{fc2}\|_{2,\mu}^2}.$$
 Then we move to layer \emph{fc2} and compute 
\begin{flalign*}
\delta'\bW_{fc2} = \delta \bW_{fc2} / m / s_{fc2} =\frac{\|\bW_{fc2}\|_{2,\mu}^2}{\|\bW_{fc3}^T\|_{2,\mu}^2} \cdot \delta \bW,
\end{flalign*} 
and update the backward factor $$m \leftarrow m\cdot \delta a_j/\delta' a_j  =\frac{(\|\bW_{fc3}^T\|_{2,\mu}^2 \cdot \|\bW_{fc2}^T\|_{2,\mu}^2)}{\|\bW_{fc2}\|_{2,\mu}^2}.$$
We continue doing this till the bottom layer.  Further details are provided in supplemental material.

\begin{figure}[ht]
\vskip 0.2in
\begin{center}
\centerline{\includegraphics[height=0.9\columnwidth]{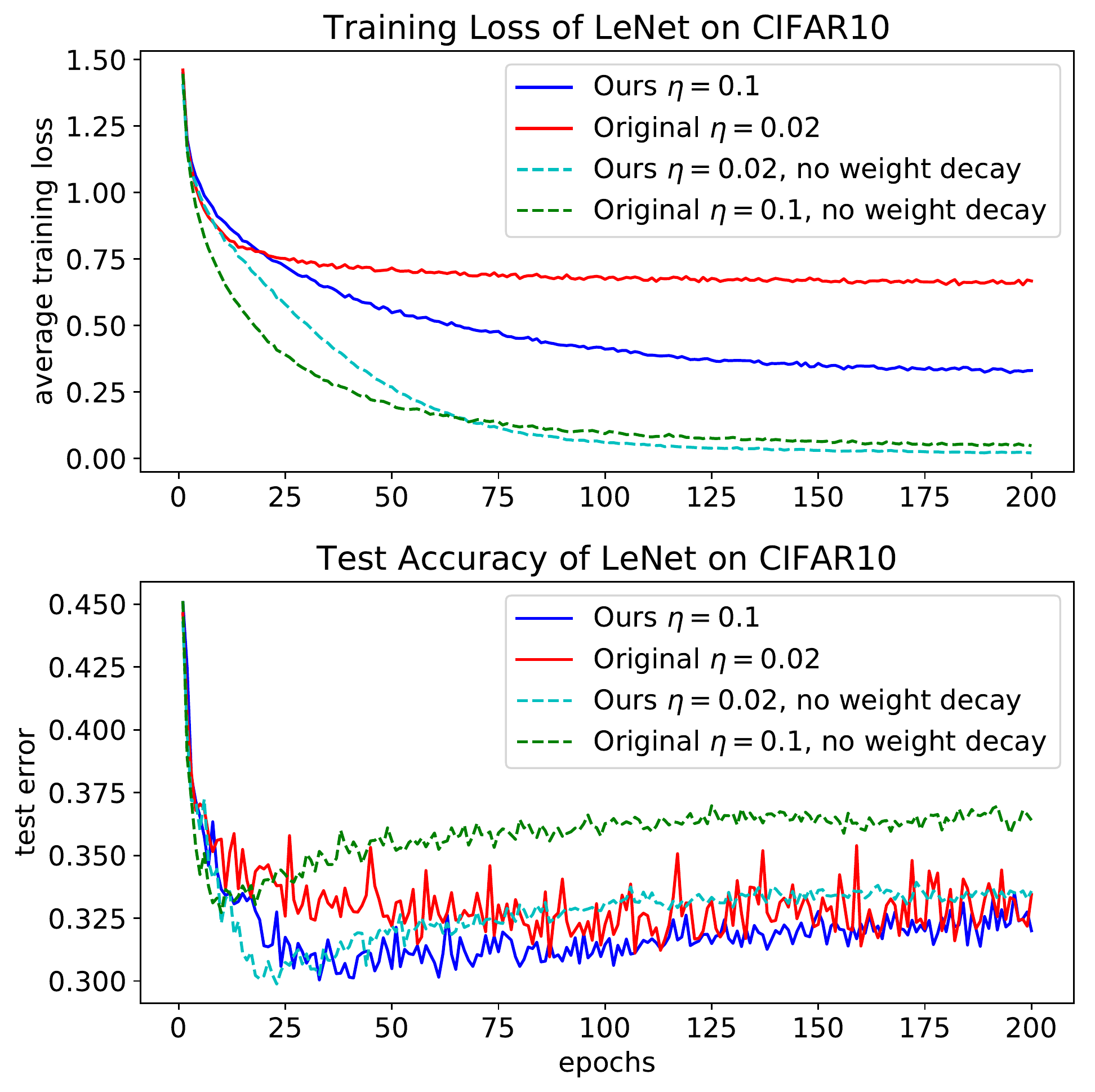}}
\caption{Performance comparison between regular momentum-SGD and our algorithm on CIFAR-10 classification using LeNet with fixed learning rate.}
\label{fig:lenetCifar10}
\end{center}
\vskip -0.2in
\end{figure}

We next train LeNet with BN to classify the  CIFAR-10 dataset \cite{cifar}. CIFAR-10 is composed of 60,000 $32\times 32$color images in 10 classes, with 6000 images per class. There are 50,000 training images and 10,000 test images. We want to compare the training procedure and test accuracy of the classical SGD based on BP and our algorithm. There are many hyper-parameters/hyper-routines that would affect the learning curve significantly, and we try our best to make a fair comparison. In the first experiment, we use only CIFAR-10 dataset without augmentation and fix momentum to be $0.9$ for both algorithm, two models start from the same initial point and pass the same mini batch of data, where the mini-batch size is 128.  Global learning rates are chosen to perform best in terms of test accuracy from a pool of five candidates\footnote{The pool for regular SGD is $\{0.02, 0.05, 0.1, 0.2, 0.5\}$ and the pool for our algorithm is $\{0.005, 0.01, 0.02, 0.05, 0.1\}$.}. We choose global learning rate  $\eta=0.1$ for SGD  and $\eta=0.02$ for our algorithm. We fix the global learning rate through the training process and train for 200 epochs. A weight-decay term (0.0005) is optional for both methods. We note that the loss in the training curve does not include the weight-decay term in Figure.~\ref{fig:lenetCifar10} no matter whether the weight-decay is used. 

From Figure.~\ref{fig:lenetCifar10}, we can see that without weight decay both algorithms can drive the training loss to zero but test accuracy of the regular SGD turns worse after very few epochs (15)  and the model becomes overfitting from then on. In comparison, our algorithm achieves a much lower test error than the regular SGD and there is a considerable margin between our algorithm and the regular SGD even in the overfitting phase. 
On the other hand, with weight decay both training losses do not converge to zero any more but our algorithm achieves much lower training loss than the regular SGD. Weight decay improves the final test accuracy but does not improve the lowest test error during the training period of our algorithm. The weight obtained by our algorithm have rather large magnitude (the network does not explode as the batch normalization stabilize the forward propagation). We believe our algorithm combats overfitting differently from the weight decay and could provide another way to improve generalization in certain setting that cannot achieved by weight decay.

\section{Experiments}
In this section, we evaluate the proposed algorithm for image classification tasks with two datasets: CIFAR-10 \citep{cifar}, CIFAR-100 \citep{cifar}.  CIFAR-10 has been introduced in Section~\ref{subsec:lenet}. The CIFAR-100 dataset is similar to the CIFAR-10, except it has 100 classes with 600 images per class and there are 500 training images and 100 testing images per class. We train VGG networks \citep{simonyan2014very} to classify these two datasets because they are widely used baselines for image classification tasks and they are of feedforward architecture.  We modify the VGG nets by keeping the last fully connected layers and removing the intermediate two fully connected layers and all the biases \footnote{We find this does not hurt accuracy for CIFAR dataset and shortens training time due to fewer parameters.}.  We equip each intermediate layer of the VGG nets with batch normalization transformation right before the activation function and the batch normalization has no trainable parameters.

Differently from the setting in Section \ref{subsec:lenet}, here we train VGG nets by using the randomly augmented CIFAR-10 and CIFAR-100 datasets (random flip and rotation) as such big models get overfitting to the datasets rapidly. We note that  augmenting the dataset the training does not gain much benefit directly. We need to decay the learning rate to learn effectively with data augmentation.  In order to compare fairly, we apply the same learning rate scheduling strategy to all algorithms: multiplying the learning rate by a factor $0.2$ every 60 epochs. 

%
%

\subsection{Baseline Algorithms}
We introduce several baseline algorithms and their settings. 

The base algorithm is the regular SGD with \emph{Nesterov momentum} $0.9$. The learning rate is set to be $\eta=0.1$. 

The second baseline algorithm is LSALR \cite{singh2015layer} which uses $\eta \cdot (1+\log (1+1/\|\delta \bW_l\|_2))$ as the learning rate for the layer $l$. The global learning rate is set to be $\eta=0.1$, which achieves best performance comparing from a pool of candidates $\{0.006, 0.05, 0.1, 0.2, 1\}$, and is different from the suggestion ($0.006$) in the original paper. 

The third baseline algorithm is LARS \cite{you2017large} which uses $\eta \cdot \frac{\|\bW_l\|_2}{\|\delta \bW_l\|_2}$ as the learning rate for layer $l$.  In our experiment, we use the global learning rate $\eta = 2$ for LARS, which achieves best performance from a pool of $\{1, 2, 5, 10\}$. 
%

Noting that all these layer-wise adaptive algorithms modify the regular layer gradient computed through BP,  we equip \emph{Nesterov momentum} $0.9$ on them  in the experiment.  For baseline algorithms, we apply \emph{weight decay} with coefficient \emph{5e-4} if without specific description. 

\subsection{Result}

\begin{figure}[ht]
\vskip 0.2in
\begin{center}
\centerline{\includegraphics[height=0.9\columnwidth]{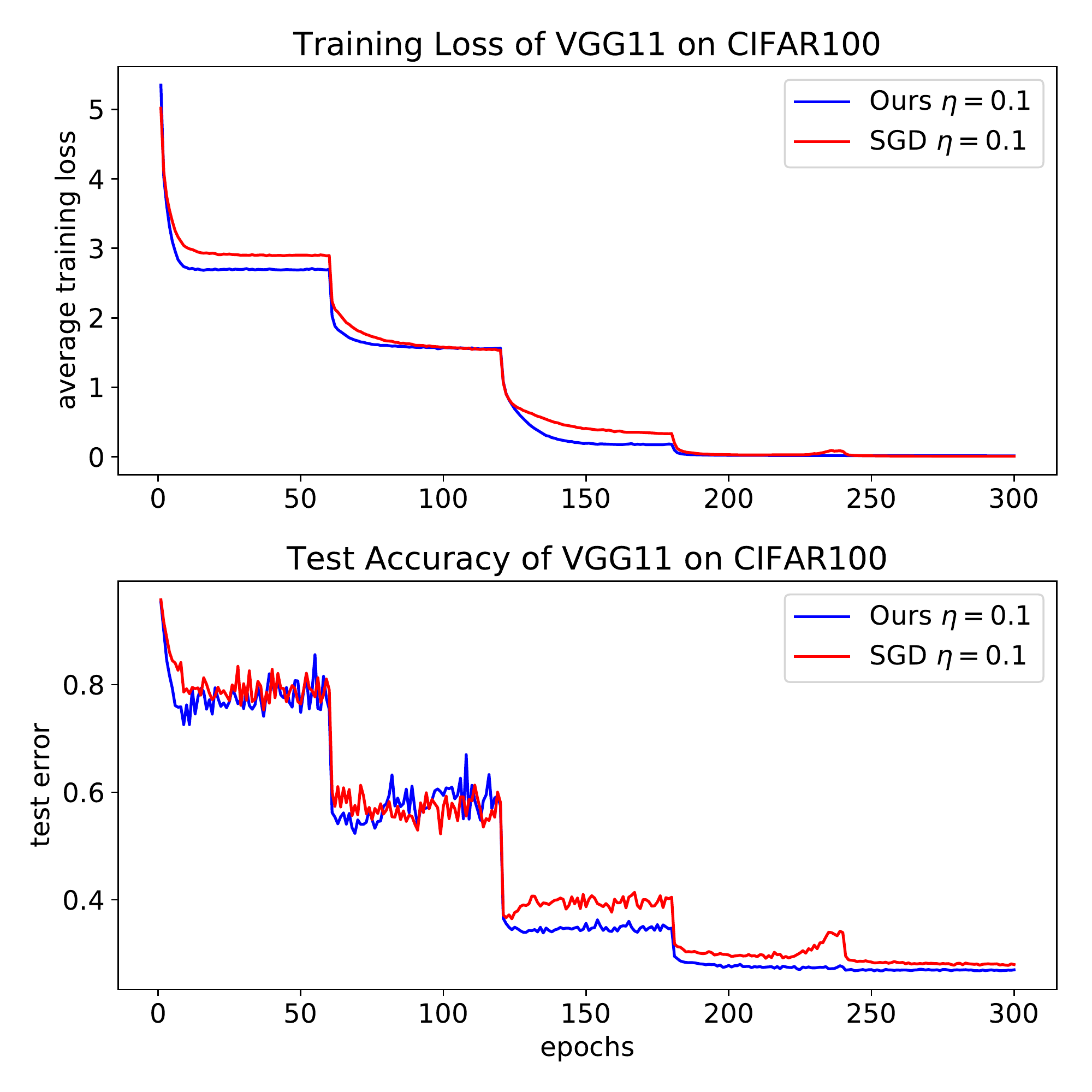}}
\caption{Performance comparison between regular momentum-SGD and our algorithm on CIFAR-10 classification using LeNet with data augmentation and learning rate scheduling.}
\label{fig:vgg11cifar100}
\end{center}
\vskip -0.2in
\end{figure}

We first compare the learning curves between the our algorithm and the vanilla SGD on training VGG11 with CIFAR-100. We apply Nesterov momentum 0.9 and weight decay coefficient 5e-3 for both algorithms. Similarly to Section \ref{subsec:lenet}, two algorithms start from the same initialization and pass the same batches of data.  We set the same learning rate $\eta=0.1$ for both algorithms. Both algorithms are run 300 epochs. We plot the learning curves in Figure~\ref{fig:vgg11cifar100}.
From Figure~\ref{fig:vgg11cifar100}, we can see that the learning curves of our algorithm and SGD have similar trend: curves jump at each learning rate decay. This is predictable as our algorithm only modifies the magnitude of the layer's gradient as a whole and does not involve any further information (second order information) and moreover we use the same hyper-parameters and the same learning rate scheduling strategy for both algorithms. Scrutinizing more closely, we can see our training loss curve is almost always lower than SGD's and our test error fluctuates heavier initially but ends with a considerably lower number.

Next we present the test result of different VGG nets for classification of CIFAR-100 in Table~\ref{tab:cifar100}. For this group of experiments, we use global learning rate $\eta=0.1$ and weight decay coefficient \emph{5e-3} for our algorithm. Our algorithm achieves higher test accuracy over its competitors on all four VGG models with margins.
\begin{table}[t]
\caption{Classification accuracies for CIFAR-100.}
\label{tab:cifar100}
\vskip 0.15in
\begin{center}
\begin{small}
\begin{sc}
\begin{tabular}{lcccr}
\toprule
Model & VGG11 & VGG13 & VGG16 & VGG19 \\
\midrule
SGD   & 71.47& 74.01&72.86 &71.35 \\
LARS& 67.26&70.21&69.90 &69.52\\
LSALR   & 70.75& 73.74&72.56 &70.76\\
Ours  	& \textbf{73.39}& \textbf{75.32} & \textbf{74.46} &\textbf{72.90} \\
\bottomrule
\end{tabular}
\end{sc}
\end{small}
\end{center}
\vskip -0.1in
\end{table}

We then present the test accuracy result of different VGG nets for classification of CIFAR-10 in Table~\ref{tab:cifar10}. The numbers in the table are the best of five independent trials of each algorithm. We use learning rate \emph{2e-3} and weight decay coefficient \emph{1e-4} for this group of experiments. We use a different learning rate from the case of CIFAR-100. The reason is that for training CIFAR-10 the last layer of VGG nets has 10 output and then the backpropagated values shall be multiplied by $\frac{512}{10}$ if following the back-matching propagation rule. 
 Such an imbalanced mapping layer makes our algorithm behave aggressively. Hence we reduce the learning rate to \emph{2e-3} for consistent result. We can see that our algorithm achieves higher test accuracy as its competitors on almost all  VGG models with various margins.

\begin{table}[t]
\caption{Classification accuracies for CIFAR-10.}
\label{tab:cifar10}
\vskip 0.15in
\begin{center}
\begin{small}
\begin{sc}
\begin{tabular}{lcccr}
\toprule
Model & VGG11 & VGG13 & VGG16 & VGG19 \\
\midrule
SGD   & 92.63&93.90&93.72 &93.66\\
LARS &91.81& 93.20&94.00 &93.48\\
LSALR    & 92.58& 93.81&94.00 &93.46\\
Ours 	& 92.69& \textbf{94.08} & \textbf{94.22}&\textbf{93.98}\\
\bottomrule
\end{tabular}
\end{sc}
\end{small}
\end{center}
\vskip -0.1in
\end{table}

%

\section{Conclusion and Discussion}
In this paper we present the back-matching propagation which provides a principled way of computing the backpropagated values on the weight parameter and on the input, which try to match the error guiding signal on the output as accurately as possible. To utilize the idea of back-matching propagation in training large neural networks efficiently, we make several approximations based on intuitive understanding and reduce the back-matching propagation to the regular BP with a layer-wise adaptive learning rate strategy. It it easy to implement within current machine learning frameworks that are equipped with auto-differentiation. We test our algorithm in training feedforward neural networks and achieve favorable result over SGD.

There are several future directions along with this work. In our derivation of the Algorithm \ref{alg:bmp}, we assume that each neuron updates its values independently from others in the same layer. This is a strong assumption and may produce inaccuracy on computing the backpropagated values across layers. 
Thus one future direction is to modify the back-matching propagation by considering the co-update of neurons in the same layer, which is closely related to Riemannian algorithms \cite{ollivier2015riemannian} that have been introduced but not widely used because of their complexity. Moreover, applying the idea of back-matching propagation to other  architectures like residual networks  and recurrent neural networks is also under consideration.


\begin{thebibliography}{30}
\providecommand{\natexlab}[1]{#1}
\providecommand{\url}[1]{\texttt{#1}}
\expandafter\ifx\csname urlstyle\endcsname\relax
  \providecommand{\doi}[1]{doi: #1}\else
  \providecommand{\doi}{doi: \begingroup \urlstyle{rm}\Url}\fi

\bibitem[Abadi et~al.(2016)Abadi, Barham, Chen, Chen, Davis, Dean, Devin,
  Ghemawat, Irving, Isard, et~al.]{tensorflow}
Abadi, Mart{\'\i}n, Barham, Paul, Chen, Jianmin, Chen, Zhifeng, Davis, Andy,
  Dean, Jeffrey, Devin, Matthieu, Ghemawat, Sanjay, Irving, Geoffrey, Isard,
  Michael, et~al.
\newblock {T}ensor{F}low: A system for large-scale machine learning.
\newblock In \emph{OSDI}, volume~16, pp.\  265--283, 2016.

\bibitem[Amari(1998)]{amari1998natural}
Amari, Shun-Ichi.
\newblock Natural gradient works efficiently in learning.
\newblock \emph{Neural computation}, 10\penalty0 (2):\penalty0 251--276, 1998.

\bibitem[Ba et~al.(2016)Ba, Kiros, and Hinton]{ba2016layer}
Ba, Jimmy~Lei, Kiros, Jamie~Ryan, and Hinton, Geoffrey~E.
\newblock Layer normalization.
\newblock \emph{arXiv preprint arXiv:1607.06450}, 2016.

\bibitem[Bastien et~al.(2012)Bastien, Lamblin, Pascanu, Bergstra, Goodfellow,
  Bergeron, Bouchard, Warde-Farley, and Bengio]{theano}
Bastien, Fr{\'e}d{\'e}ric, Lamblin, Pascal, Pascanu, Razvan, Bergstra, James,
  Goodfellow, Ian, Bergeron, Arnaud, Bouchard, Nicolas, Warde-Farley, David,
  and Bengio, Yoshua.
\newblock Theano: new features and speed improvements.
\newblock \emph{arXiv preprint arXiv:1211.5590}, 2012.

\bibitem[Duchi et~al.(2011)Duchi, Hazan, and Singer]{duchi2011adaptive}
Duchi, John, Hazan, Elad, and Singer, Yoram.
\newblock Adaptive subgradient methods for online learning and stochastic
  optimization.
\newblock \emph{Journal of Machine Learning Research}, 12\penalty0
  (Jul):\penalty0 2121--2159, 2011.

\bibitem[Grosse \& Martens(2016)Grosse and Martens]{grosse2016kronecker}
Grosse, Roger and Martens, James.
\newblock A {Kronecker}-factored approximate {F}isher matrix for convolution
  layers.
\newblock In \emph{International Conference on Machine Learning (ICML)}, 2016.

\bibitem[He et~al.(2016)He, Zhang, Ren, and Sun]{he2016deep}
He, Kaiming, Zhang, Xiangyu, Ren, Shaoqing, and Sun, Jian.
\newblock Deep residual learning for image recognition.
\newblock In \emph{The IEEE Conference on Computer Vision and Pattern
  Recognition (CVPR)}, June 2016.

\bibitem[Hinton et~al.(2012)Hinton, Deng, Yu, Dahl, Mohamed, Jaitly, Senior,
  Vanhoucke, Nguyen, Sainath, et~al.]{hinton2012deep}
Hinton, Geoffrey, Deng, Li, Yu, Dong, Dahl, George~E, Mohamed, Abdel-rahman,
  Jaitly, Navdeep, Senior, Andrew, Vanhoucke, Vincent, Nguyen, Patrick,
  Sainath, Tara~N, et~al.
\newblock Deep neural networks for acoustic modeling in speech recognition: The
  shared views of four research groups.
\newblock \emph{IEEE Signal Processing Magazine}, 29\penalty0 (6):\penalty0
  82--97, 2012.

\bibitem[Hochreiter \& Schmidhuber(1997)Hochreiter and
  Schmidhuber]{hochreiter1997long}
Hochreiter, Sepp and Schmidhuber, J{\"u}rgen.
\newblock Long short-term memory.
\newblock \emph{Neural Computation}, 9\penalty0 (8):\penalty0 1735--1780, 1997.

\bibitem[Ioffe \& Szegedy(2015)Ioffe and Szegedy]{ioffe2015batch}
Ioffe, Sergey and Szegedy, Christian.
\newblock Batch normalization: Accelerating deep network training by reducing
  internal covariate shift.
\newblock In \emph{International Conference on Machine Learning (ICML)}, pp.\
  448--456, 2015.

\bibitem[Kingma \& Ba(2014)Kingma and Ba]{kingma2014adam}
Kingma, Diederik and Ba, Jimmy.
\newblock Adam: A method for stochastic optimization.
\newblock \emph{arXiv preprint arXiv:1412.6980}, 2014.

\bibitem[Krizhevsky \& Hinton(2009)Krizhevsky and Hinton]{cifar}
Krizhevsky, Alex and Hinton, Geoffrey.
\newblock Learning multiple layers of features from tiny images.
\newblock 2009.

\bibitem[Lafond et~al.(2017)Lafond, Vasilache, and Bottou]{lafond2017diagonal}
Lafond, Jean, Vasilache, Nicolas, and Bottou, L{\'e}on.
\newblock Diagonal rescaling for neural networks.
\newblock \emph{arXiv preprint arXiv:1705.09319}, 2017.

\bibitem[LeCun et~al.(1998{\natexlab{a}})LeCun, Bottou, Bengio, and
  Haffner]{lecun1998gradient}
LeCun, Yann, Bottou, L{\'e}on, Bengio, Yoshua, and Haffner, Patrick.
\newblock Gradient-based learning applied to document recognition.
\newblock \emph{Proceedings of the IEEE}, 86\penalty0 (11):\penalty0
  2278--2324, 1998{\natexlab{a}}.

\bibitem[LeCun et~al.(2015)LeCun, Bengio, and Hinton]{lecun2015deep}
LeCun, Yann, Bengio, Yoshua, and Hinton, Geoffrey.
\newblock Deep learning.
\newblock \emph{Nature}, 521\penalty0 (7553):\penalty0 436, 2015.

\bibitem[LeCun et~al.(1998{\natexlab{b}})LeCun, Bottou, Orr, and
  M{\"u}ller]{lecun1998efficient}
LeCun, Yann~A, Bottou, L{\'e}on, Orr, Genevieve~B, and M{\"u}ller,
  Klaus-Robert.
\newblock Efficient backprop.
\newblock In \emph{Neural networks: Tricks of the trade}. Springer,
  1998{\natexlab{b}}.

\bibitem[Marceau-Caron \& Ollivier(2016)Marceau-Caron and
  Ollivier]{marceau2016practical}
Marceau-Caron, Ga{\'e}tan and Ollivier, Yann.
\newblock Practical {R}iemannian neural networks.
\newblock \emph{arXiv preprint arXiv:1602.08007}, 2016.

\bibitem[Martens \& Grosse(2015)Martens and Grosse]{martens2015optimizing}
Martens, James and Grosse, Roger.
\newblock Optimizing neural networks with {Kronecker}-factored approximate
  curvature.
\newblock In \emph{International Conference on Machine Learning (ICML)}, pp.\
  2408--2417, 2015.

\bibitem[Nesterov(2013)]{nesterov2013introductory}
Nesterov, Yurii.
\newblock \emph{Introductory lectures on convex optimization: A basic course},
  volume~87.
\newblock Springer Science \& Business Media, 2013.

\bibitem[Ollivier(2015)]{ollivier2015riemannian}
Ollivier, Yann.
\newblock {R}iemannian metrics for neural networks {I}: feedforward networks.
\newblock \emph{Information and Inference: A Journal of the IMA}, 4\penalty0
  (2):\penalty0 108--153, 2015.

\bibitem[Paszke et~al.(2017)Paszke, Gross, Chintala, Chanan, Yang, DeVito, Lin,
  Desmaison, Antiga, and Lerer]{pytorch}
Paszke, Adam, Gross, Sam, Chintala, Soumith, Chanan, Gregory, Yang, Edward,
  DeVito, Zachary, Lin, Zeming, Desmaison, Alban, Antiga, Luca, and Lerer,
  Adam.
\newblock Automatic differentiation in {P}y{T}orch.
\newblock 2017.

\bibitem[Qian(1999)]{qian1999momentum}
Qian, Ning.
\newblock On the momentum term in gradient descent learning algorithms.
\newblock \emph{Neural networks}, 12\penalty0 (1):\penalty0 145--151, 1999.

\bibitem[Rumelhart et~al.(1986)Rumelhart, Hinton, and
  Williams]{rumelhart1986learning}
Rumelhart, David~E, Hinton, Geoffrey~E, and Williams, Ronald~J.
\newblock Learning representations by back-propagating errors.
\newblock \emph{nature}, 323\penalty0 (6088):\penalty0 533, 1986.

\bibitem[Seide \& Agarwal(2016)Seide and Agarwal]{CNTK}
Seide, Frank and Agarwal, Amit.
\newblock {CNTK}: {M}icrosoft's open-source deep-learning toolkit.
\newblock In \emph{Proceedings of the 22nd ACM SIGKDD International Conference
  on Knowledge Discovery and Data Mining}, pp.\  2135--2135. ACM, 2016.

\bibitem[Simonyan \& Zisserman(2015)Simonyan and Zisserman]{simonyan2014very}
Simonyan, Karen and Zisserman, Andrew.
\newblock Very deep convolutional networks for large-scale image recognition.
\newblock In \emph{ICLR}, 2015.

\bibitem[Singh et~al.(2015)Singh, De, Zhang, Goldstein, and
  Taylor]{singh2015layer}
Singh, B., De, S., Zhang, Y., Goldstein, T., and Taylor, G.
\newblock Layer-specific adaptive learning rates for deep networks.
\newblock In \emph{IEEE 14th International Conference on Machine Learning and
  Applications (ICMLA)}, pp.\  364--368, Dec 2015.

\bibitem[Sutskever et~al.(2013)Sutskever, Martens, Dahl, and
  Hinton]{sutskever2013importance}
Sutskever, Ilya, Martens, James, Dahl, George~E, and Hinton, Geoffrey~E.
\newblock On the importance of initialization and momentum in deep learning.
\newblock \emph{International Conference on Machine Learning (ICML)},
  28:\penalty0 1139--1147, 2013.

\bibitem[Ye et~al.(2017)Ye, Yang, Fermuller, and Aloimonos]{ye2017importance}
Ye, Chengxi, Yang, Yezhou, Fermuller, Cornelia, and Aloimonos, Yiannis.
\newblock On the importance of consistency in training deep neural networks.
\newblock \emph{arXiv preprint arXiv:1708.00631}, 2017.

\bibitem[You et~al.(2017)You, Gitman, and Ginsburg]{you2017large}
You, Yang, Gitman, Igor, and Ginsburg, Boris.
\newblock Large batch training of convolutional networks.
\newblock \emph{arXiv preprint arXiv:1708.03888v3}, 2017.

\bibitem[Zhang et~al.(2017)Zhang, Xiong, Bradbury, and Socher]{zhang2017block}
Zhang, Huishuai, Xiong, Caiming, Bradbury, James, and Socher, Richard.
\newblock Block-diagonal hessian-free optimization for training neural
  networks.
\newblock \emph{arXiv preprint arXiv:1712.07296}, 2017.

\end{thebibliography}
\bibliographystyle{icml2018}

\clearpage

\onecolumn
\appendix

\noindent {\Large \textbf{Supplementary Material}}

We walk through the approximated back-matching propagation of the LeNet and show how each layer's weight should be changed ($\delta'\bW$) under the rule of the approximated back-matching propagation.
Following the procedure of Algorithm \ref{alg:bmp}, we have the following initial value: $m=1, s_{fc3} = 1, s_{fc2} = 1, s_{fc1} = 1, s_{cv2} = 25, s_{cv1} = 196$. Given the loss $\ell(x)$, we can compute the normal gradient on each weight parameter through BP, which are denoted as $\delta \bW$ with subscript of the layer name. We start from the top layer \emph{fc3} and compute 
\begin{flalign*}
\delta'\bW_{fc3} = \delta \bW_{fc3} / m / s_{fc3} = \delta \bW
\end{flalign*} 
and update $$m \leftarrow m\cdot \delta a_j/\delta' a_j = \|\bW_{fc3}^T\|_{2,\mu}^2.$$
Since the ReLU activation does not contain parameter and does not change the backward factor $m$, then we move to the BN layer. Since our BN layer does not have parameter, we only have to update the backward factor $$m \leftarrow m\cdot \delta a_j/\delta' a_j  =\|\bW_{fc3}^T\|_{2,\mu}^2/\|\bW_{fc2}\|_{2,\mu}^2.$$
 Then we move to layer \emph{fc2} and compute 
\begin{flalign*}
\delta'\bW_{fc2} = \delta \bW_{fc2} / m / s_{fc2} =\frac{\|\bW_{fc2}\|_{2,\mu}^2}{\|\bW_{fc3}^T\|_{2,\mu}^2} \cdot \delta \bW,
\end{flalign*} 
and update the backward factor $$m \leftarrow m\cdot \delta a_j/\delta' a_j  =(\|\bW_{fc3}^T\|_{2,\mu}^2 \cdot \|\bW_{fc2}^T\|_{2,\mu}^2)/\|\bW_{fc2}\|_{2,\mu}^2.$$ Then after another BN layer, the backward factor becomes $$m \leftarrow m\cdot \delta a_j/\delta' a_j  =(\|\bW_{fc3}^T\|_{2,\mu}^2 \cdot \|\bW_{fc2}^T\|_{2,\mu}^2)/(\|\bW_{fc2}\|_{2,\mu}^2 \cdot \|\bW_{fc1}\|_{2,\mu}^2).$$ Then we move to layer \emph{fc1} and compute
\begin{flalign*}
\delta'\bW_{fc1} = \delta \bW_{fc1} / m / s_{fc1} =\frac{(\|\bW_{fc2}\|_{2,\mu}^2 \cdot \|\bW_{fc1}\|_{2,\mu}^2)}{(\|\bW_{fc3}^T\|_{2,\mu}^2 \cdot \|\bW_{fc2}^T\|_{2,\mu}^2)} \cdot \delta \bW,
\end{flalign*}
and update the backward factor $$m \leftarrow m\cdot \delta a_j/\delta' a_j  =(\|\bW_{fc3}^T\|_{2,\mu}^2 \cdot \|\bW_{fc2}^T\|_{2,\mu}^2 \cdot  \|\bW_{fc1}^T\|_{2,\mu}^2)/(\|\bW_{fc2}\|_{2,\mu}^2 \cdot \|\bW_{fc1}\|_{2,\mu}^2).$$ 
After another BN layer, the backward factor becomes $$m \leftarrow m\cdot \delta a_j/\delta' a_j  =\frac{(\|\bW_{fc3}^T\|_{2,\mu}^2 \cdot \|\bW_{fc2}^T\|_{2,\mu}^2 \cdot  \|\bW_{fc1}^T\|_{2,\mu}^2)}{(\|\bW_{fc2}\|_{2,\mu}^2 \cdot \|\bW_{fc1}\|_{2,\mu}^2\cdot \|\bW_{cv2, row}\|_{2,\mu}^2)}.$$ 
Then we move the convolutional layer \emph{cv2} and compute 
\begin{flalign*}
\delta'&\bW_{cv2} = \delta \bW_{cv2} / m / s_{cv2} \\
& =\frac{(\|\bW_{fc2}\|_{2,\mu}^2 \cdot \|\bW_{fc1}\|_{2,\mu}^2\cdot \|\bW_{cv2, row}\|_{2,\mu}^2)}{25\cdot(\|\bW_{fc3}^T\|_{2,\mu}^2 \cdot \|\bW_{fc2}^T\|_{2,\mu}^2 \cdot  \|\bW_{fc1}^T\|_{2,\mu}^2)} \cdot \delta \bW_{cv2}
\end{flalign*}
and update the backward factor $$m \leftarrow m\cdot \delta a_j/\delta' a_j  =\frac{(\|\bW_{fc3}^T\|_{2,\mu}^2 \cdot \|\bW_{fc2}^T\|_{2,\mu}^2 \cdot  \|\bW_{fc1}^T\|_{2,\mu}^2\cdot \|\bW_{cv2, col}\|_{2,\mu}^2/(196/25))}{(\|\bW_{fc2}\|_{2,\mu}^2 \cdot \|\bW_{fc1}\|_{2,\mu}^2\cdot \|\bW_{cv2, row}\|_{2,\mu}^2)}.$$ 
After another BN layer, the backward factor becomes
$$m \leftarrow m\cdot \delta a_j/\delta' a_j  =\frac{(\|\bW_{fc3}^T\|_{2,\mu}^2 \cdot \|\bW_{fc2}^T\|_{2,\mu}^2 \cdot  \|\bW_{fc1}^T\|_{2,\mu}^2\cdot \|\bW_{cv2, col}\|_{2,\mu}^2/(196/25))}{(\|\bW_{fc2}\|_{2,\mu}^2 \cdot \|\bW_{fc1}\|_{2,\mu}^2\cdot \|\bW_{cv2, row}\|_{2,\mu}^2\cdot  \|\bW_{cv1, row}\|_{2,\mu}^2)}.$$ 
Finally, we move to the bottom convolutional layer \emph{cv1} and compute
\begin{flalign*}
\delta'&\bW_{cv1} = \delta \bW_{cv1} / m / s_{cv1} \\
& =\frac{(\|\bW_{fc2}\|_{2,\mu}^2 \cdot \|\bW_{fc1}\|_{2,\mu}^2\cdot \|\bW_{cv2, row}\|_{2,\mu}^2\cdot  \|\bW_{cv1, row}\|_{2,\mu}^2)}{196(\|\bW_{fc3}^T\|_{2,\mu}^2 \cdot \|\bW_{fc2}^T\|_{2,\mu}^2 \cdot  \|\bW_{fc1}^T\|_{2,\mu}^2\cdot \|\bW_{cv2, col}\|_{2,\mu}^2/(196/25))} \cdot \delta \bW_{cv1} \\
& =\frac{(\|\bW_{fc2}\|_{2,\mu}^2 \cdot \|\bW_{fc1}\|_{2,\mu}^2\cdot \|\bW_{cv2, row}\|_{2,\mu}^2\cdot  \|\bW_{cv1, row}\|_{2,\mu}^2)}{25(\|\bW_{fc3}^T\|_{2,\mu}^2 \cdot \|\bW_{fc2}^T\|_{2,\mu}^2 \cdot  \|\bW_{fc1}^T\|_{2,\mu}^2\cdot \|\bW_{cv2, col}\|_{2,\mu}^2)} \cdot \delta \bW_{cv1}
\end{flalign*}

\end{document}